# Detection and emergence

Eric Bonabeau[1,3], Jean-Louis Dessalles[2,4]

(1) Santa Fe Institute, 1399 Hyde Park Road, Santa Fe, NM 87501, USA
(2) Département Informatique, ENST, 46 rue Barrault, 75013 Paris
(3) email: bonabeau@santafe.edu
(4) email: dessalles@ enst.fr

### Abstract

Two different conceptions of emergence are reconciled as two instances of the phenomenon of detection. In the process of comparing these two conceptions, we find that the notions of complexity and detection allow us to form a unified definition of emergence that clearly delineates the role of the observer.

### Résumé

Deux définitions différentes de l'émergence sont ici rapprochées comme étant deux cas de détection. Les notions de complexité et de détection nous permettent d'offrir une définition unique de l'émergence qui délimite clairement le rôle de l'observateur.

Mots-clés: détection, émergence, modèles, complexité, niveaux d'organisation

## 1. Introduction

Emergence is obviously an ill-defined concept, which has been given many, apparently incompatible, definitions (Bonabeau et al., 1995a). Some of them refer to levels of organization (Lewes, 1874). Emerging properties would be relevant only to the upper level, even if they are grounded in the lower level. Other definitions refer to self-organization (Varela *et al.*, 1991), to entropy changes (Kauffman, 1990), to non-linearity (Langton, 1990), to deviations from predicted behavior (Rosen, 1985, Cariani 1991) or from symmetry (Palmer, 1989). Other definitions are closely related to the concept of complexity (Bonabeau et al., 1995a, 1995b; Cariani, 1991; Kampis, 1991). Among all these definitions, some are presented as more "objective" than others which refer explicitly to an observer (Bonabeau et al., 1995b).

We propose here a conceptual framework, based on the notion of *detection*, in which all these definitions of emergence may be compared. We first consider well-accepted characterizations of emergence, which refer to levels of organization and to model changes. Then we show that emergence is related to complexity shifts. Lastly, we propose to focus on the observer, rather on the emerging system, in order to show that all characterizations of emergence are implicitly connected to the notion of detection.

## 2. Defining emergence

Emergence is foremost an intuitive phenomenon. An easy way to illustrate this idea is to consider scale changes. When walking in the streets of a European city, an observer may not detect its fractal structure, whereas an aerial or satellite photograph of the city would reveal it. Another example in the same vein involves patterns of vegetation aggregation (Dayong, 1990): the spatial pattern of desert shrubs may display a combination of small-scale regularity (generated through competition among individuals) and large-scale aggregation (due to the clustered seedling establishment resulting from limited seed dispersal). When looking at these patterns on either a large or a small scale, one will miss the global organization which requires, to be understood, that both scales are simultaneously taken into account. Classical statistics, indeed, will tend to confound small-scale regularity with large-scale aggregation. Yet another example from ecology comes from the fractal dimension of forests (Sugihara and May, 1990): the fractal dimension of deciduous forest patterns in Mississippi (Natchez Quadrangle) has been measured by plotting the logarithm of the patch perimeter against the patch area. A discontinuity is discovered in the fractal dimension, indicating a shift in dimension from $D=1.2$ to $D=1.5$, at area scales of about 60-70 ha. Since native forests in this region have recently been converted into agricultural use, this shift in dimension may result from human activities. Small forest areas (<60 ha) are indeed relatively smoother than larger areas (>70 ha). This is due to the fact that human disturbances dominate at small scales, making forest patterns smoother, while natural processes, such as geology or the distribution of soil types, continue to dominate at larger scales (Sugihara and May, 1990). If this interpretation is correct, the scale at which the dimension shift takes place should increase with the expansion of human disturbances. Once again, here, if the system is observed over an insufficient number of scales, the effect of human activities cannot be detected. Another example from neurobiology (Swindale, 1980) illustrates the same idea from a slightly different viewpoint: among the patterns of ocular dominance in the visual cortex of monkeys, stripes of right-eye dominance alternate with stripes of left-eye dominance, covering the cortex like the coat of a zebra. For an observer studying the level of synapses, these patterns cannot be detected, whereas looking at 2D cuts of the cortex over scales of the order of a few hundreds of µm allows to detect these patterns, which, as many models suggest, result from short-range activation and long-range inhibition among synapses.

In all these examples, some relevant phenomenon remains hidden if the observer looks at the wrong scales. Such phenomena *emerge* when the observer begins to consider the correct scale. It seems that "objective" notions like shifts in fractal dimension make the presence of an observer unnecessary. We will suggest that is only an illusion, and that the concept of emergence is much better captured through a sound analysis of the observer's structure, if we accept to generalize the observer's role.

A classical attempt to give a general definition of emergence consists in considering *levels*. According to this view, there is emergence whenever a "phenomenon" appears at a level $L_h$ out of constituents and processes defined and taking place at a lower level $L_l$, and when the properties of $L_h$ are "difficult" or impossible to deduce from the properties of the constituent units and processes of $L_l$. Let us name this type of emergence "emergence of higher-level structures" (EHS). We may distinguish two cases: (i) a structure appears at $L_h$ in the course of time through the natural dynamics of the system, or (ii) the "structure" was potentially there before, but it was not actualized in $L_h$ until some observational tool was available. The first case (i) refers, for example, to self-organizing systems in physics, chemistry or biology (Nicolis and Prigogine, 1989), whereas (ii) refers, for instance, to the above mentioned example of the fractal structure of a city, which becomes apparent only with appropriate tools (large-scale observations, such as satellite observations). Note that the very notions of "structure" and "phenomenon" are still quite ambiguous here.

Another conception of emergence is called emergence-relative-to-a-model (ERM) (Cariani, 1991; Rosen, 1978, 1985). For the supporters of ERM, there is emergence when a natural system's behavior can no longer be explained by a model of that system (it deviates or "bifurcates" from the predictions of the model), and requires a new model. According to Pattee (1989), theories like fractals, catastrophe theory, dissipative structures, symmetry-breaking instabilities and chaotic dynamics do not convey the full essence of emergence, esp. of biological emergence, since such formal models are unable to produce entirely novel structures. They are doomed to deviate from the behavior of the actual system. In order to keep up with the system's behavior, the modeler either has to (i) find new relationships between existing variables (syntactic emergence), or (ii) find new observables associated with the natural system which can be transformed into variables for modeling purposes (semantic emergence).

We can easily show that both approaches, EHS and ERM, implicitly involve an observer. The existence of an emerging structure in $L_h$ is only possible if it can be detected at that level thanks to some observational tool. Similarly, the connection, through observables, between the model and the emerging phenomenon looks like the relationship between the observer and the observed system. In what follows, we use the notion of complexity to better formalize this link between levels, models, observer and emergence.

## 3. Complexifying emergence

Emergence and complexity do indeed share a lot of common features. Before attempting to present a unified picture of EHS and ERM, let us briefly introduce the notion of complexity. Complexity is a relative concept, which depends both on the task at hand and on the tools available to achieve this task (Bennett, 1990; Bonabeau, 1993; Crutchfield and Young, 1990; Grassberger, 1989; Kampis, 1991). A first idea of complexity is given by entropy, which measures the lack of information of a given observer about a system. A more elaborate approach to complexity consists in evaluating the length of the shortest algorithm which is able to reconstruct the observed system (Kolmogorov-Chaitin-Solomonoff algorithmic complexity). The notion of complexity can be refined to take into account the internal structure perceived in the system. Given a set of tools considered as structuring elements, there are some aspects of the system that can be explained (or compressed) by means of these tools, and there are some aspects of the object that cannot be understood using these tools. Such tools may be Turing machines, as in the standard definition of algorithmic complexity, or more generally any device able to give a concise description of some part of the observed system. The relative complexity $C(S/T)$ of a system $S$ with respect to a set of tools $T$ can be interpreted as "the difficulty of decomposing $S$ when $T$ is used as a structuring element (*i.e.* what is explainable by $T$) + what is not decomposable by means of this structuring element (*i.e.* what is not explainable by $T$ and remains to be explained otherwise)".

Emergence can be advantageously described in this kind of framework. Like complexity, it is not an absolute notion. It is relative to the tools that one may use to observe the system to be characterized. Emergence can then be defined with respect to the same tools used to define the complexity of a system. It occurs when an object or phenomenon cannot be detected or understood with a given set of tools but can be detected or understood by allowing some additional tools. For some reason (dynamic evolution of the system or changes in the set of observational tools) a new apprehension of the system becomes possible that offers a shorter overall description, and hence a smaller relative complexity. *Emergence is thus associated with a decrease of the relative complexity*. For instance, when observational tools are characterized by a perceptive "diameter", as was the case in our example about the perception of fractal structures, emerging structures correspond to shifts in relative complexity which, in this case, takes the form of $K_d$-complexity (Chaitin, 1979).

We will examine now how this unifying description applies to EHS and ERM. This will require the introduction of the notion of detection.

## 4. Detecting emergence

The notion of detector is very basic: any device which gives a binary response to its input. Intuitively, since emergence is an all-or-nothing phenomenon, it must correspond to some detector becoming active. This very simple idea has interesting consequences for the characterization of emergence.

It is possible to define the relative complexity of a system with respect to a set of detectors. The relative complexity of a system observed through a set of detectors at a given moment will be measured by the complexity of active detectors + the complexity of the relations between these detectors. More precisely, the relative complexity should be written $C(S/D,T)$, where $D$ is a set of detectors and $T$ a set of available tools that allow to compute a description of structures detected through $D$. This constitutes a new formulation of the notion of relative complexity, the only difference being that the distinction between observational tools and description tools is now made explicit.

We can now characterize emergence. Let us consider that a given detector, $D_k$, becomes active at time $t+\Delta t$. At time $t$, $D_k$ is inactive. Taking $D_k$ into account at time $t$ does not change the relative complexity:

$$C_t(S/T, D_1, ..., D_{k-1}, D_k) = C_t(S/T, D_1, ..., D_{k-1}) \qquad (1)$$

At time $t+\Delta t$, $D_k$ becomes activated. Normally, an increase of complexity should ensue, since the description of the system performed with $T$ becomes *a priori* longer. Emergence occurs when the contrary happens:

$$C_{t+\Delta t}(S/T, D_1, ..., D_{k-1}, D_k) < C_t(S/T, D_1, ..., D_{k-1}) = C_t \qquad (2)$$

This is possible if the activity of some of the $D_i$ is redundant with the activity of $D_k$. The $D_i$ can thus be omitted in the description, and the result may be a smaller relative complexity.

We can now reconsider EHS and ERM in this framework. Levels of organization can be described as resulting from a hierarchical structure of the set of detectors (Bonabeau et al., 1995b). A detector of level $n$ receives inputs from detectors of level $n-1$ only. To borrow an example from linguistics, the ontological hierarchy sound–phoneme–word–phrase–sentence can be shown to result from a set of appropriate linguistic devices that we can describe as nested detectors. We detect the occurrence of a sentence after having detected its phrasal constituents, and not directly from the presence of words nor from its bare acoustic form. Many perceptual or computational devices exhibit this hierarchical detecting structure. When a detector becomes active in such a hierarchy, the active detectors from the lower level that are connected to it can be omitted from the description. When we recognize a word, we can forget the phonemes that compose it and that were detected in the first place (this is only an approximation when the input is noisy: erroneous phonemes should remain in an exhaustive description of the linguistic structure).

EHS fits naturally in our definition of emergence. If $D_k$ is a detector of level $n$, then all active detectors of level $n-1$ which are connected to $D_k$ become redundant. There are such detectors, since $D_k$ did not become active by magic. As a consequence, relative complexity automatically decreases and emergence necessarily occurs whenever an $n$-detector is activated while only ($n-1$)-detectors were previously active. The activation or availability of an $n$-detector systematically enables the building of a more compact – and therefore less complex – model that takes advantage of regularities and redundancy at the lower level. Emergence is thus a characteristic feature of detection hierarchies.

Note that emergence may occur at level $n=1$. The lower level, $L_0$ in this case, consists of analog sensors. Even between two genuine levels ($n-1$) and $n$, analog properties among ($n-1$)-detectors (*e.g.* neighborhood, in the visual system) may be involved in the computation of the $n$-detector's output. In such case, these analog properties are lost at level $n$, they cannot be recovered from the mere observation of the $n$-detector activity.

We will now consider ERM and show that it is a special case of EHS. A situation of ERM presupposes three successive periods: (T1) the behavior of the modeled system is correctly predicted; (T2) it departs from the predicted behavior (bifurcation); (T3) a new model emerges, thanks to the introduction of a new observable (semantic emergence) or of a new combinatorial ability (syntactic emergence). From a complexity viewpoint, a bifurcation leads to an increase in relative complexity, because one must back up to a lower level description in terms of available observables. In this respect, the introduction of a valid model of the system in T3 offers a higher-level description, which restores a concise description. This looks very much like EHS. To formalize this analogy, let us notice that a bifurcation is a binary event. This suggests that the model should be considered itself as a detector. As we will see, if we take this idea seriously, then several aspects of emergence become clearer and ERM and EHS can be unified.

## 5. Conceptualizing emergence

The prototypic idea of detector is a binary device receiving its input from analog sensors or from other detectors through hardwired circuitry. This is probably how low-level detectors like edge detectors are implemented in visual perception. We suggest that more complex devices involved in language understanding or in scientific modeling can legitimately be considered as detectors too.

For instance, Chomsky's grammars correspond to well-formalized detectors. They recognize languages as subsets of all possible strings of symbols. They can be implemented in automata: a finite-state automaton will detect strings of a regular language, a pushdown automaton will detect strings of a context-free language, a linear bounded automaton recognizes strings of a context-sensitive language. These automata perform an actual detection. However, detection is already present, conceptually, in the grammar itself.

Similarly, the use of spatial or temporal correlation functions can help discover, say, periodic features in space or time. The fact that these correlations are implemented in actual devices or remain mathematical abstractions is not relevant. If the mathematical model is able to predict some periodicity, it acts as a detector, a conceptual one. Kepler's laws can also appear as a conceptual detector: they can predict whether the trajectory of a cosmic object will be elliptic. The detecting abilities of these models are obvious in case of failure, what we called a bifurcation. If a combination of periodic signals fails to be periodic, or if the comet's trajectory is neither elliptic nor hyperbolic, then the model is able to "raise a flag" to signal that something goes wrong.

Such models rely on observables (arrays of coordinates, data like body masses, etc.) and compute predictions. In case of bifurcation, either new observables must be found (higher sampling rate, presence of other celestial bodies) or new computations must be made available (*e.g.* Eintein's formula instead of Newton's laws). As already mentioned, the former case may lead to semantic emergence, the latter to syntactic emergence. It is important to note that such emergence, ERM, consists in the restoration of an appropriate (conceptual) detector. The analogy with EHS is now clear: at period T2, the system is described only through the available observables, which belong to a lower level $L_l$ (continuous observables should be considered rather as sensors than as detectors). At period T3, a new valid model is available. It behaves as an activated detector of higher level $L_h$, receiving its input from $L_l$.

Syntactic ERM is nothing but the setting-up of a new high-level detector which becomes activated. Semantic ERM consists in introducing a new low-level detector or sensor (observable), which leads to the emergent activation of a high-level detector. In semantic ERM, the system's relative complexity is increased not only by the bifurcation, but also by the introduction of a further relevant variable. The dimension of the phase space increases so that the system's behavior is harder and longer to describe with the available tools. It is only thanks to the re-activation of the model that the complexity diminishes, leading to an emergent phenomenon. The mere mention of a nominal trajectory may for instance replace a whole set of previously unexplained arrays of coordinates.

By considering that scientific models play the role of conceptual detectors, we see that ERM is a special case of EHS. Conversely, the existence of complex detectors in ERM, like grammars or scientific models, that require sometimes heavy computations, draws the attention to the fact that in EHS, the computational cost of detectors has to be taken into account to evaluate complexity. The high-level detector which is responsible for the occurrence of emergence makes the description of the system simpler, by extracting various types of symmetry and redundancy. There is a cost, however: the cost of detection, which can be defined as the size of the recognition device (we do not consider the *depth* of computation here (Bennet, 1990)). By becoming aware of the existence of complex detectors, we may consider that low-level detectors too should be included in the computation of relative complexity. This may not be an easy task, since the notion of detection complexity is not always easily quantifiable.

## 6. Conclusion: emergence and cognition

By introducing the notion of detection, first to characterize organization levels, then to describe the behavior of scientific models in case of "bifurcation", we could offer an unified picture of emergence, embracing both emergence of higher-level structures (EHS) and emergence relative to a model (ERM). Emergence has been characterized by the activation of a high-level detector, with the effect of decreasing relative complexity.

This way of presenting emergence leads to an unexpected situation. Many attempts have been made in the literature to describe emergence in an "objective" way, *i.e.* in an observer-independent way. Levels of organization, phase spaces, complexity and models are thought to be more or less independent from the observer. In this paper, this picture has been completely inverted. Levels of organization are not out there, but result from the observer's perceptive structure which is organized as a detection hierarchy. Phase spaces are drawn in coordinates which are the problem's observables. These observables are the observer's low-level sensors (if continuous) or detectors (if binary). They are not intrinsic to the system itself. Similarly, complexity has been presented as observer-dependent: relative complexity makes an explicit reference to available detectors and combinatorial tools. Even models have been presented as detectors. They are not part of the system. They are conceptual tools, and as such they are observer-dependent. The model may change without affecting the system itself in any way. What remains at the end is a scenario of emergence which takes place entirely on the observer's side. This result should not be considered as negative. What we name "observer" here may indeed refer to abstract entities like computational devices, grammars or scientific models, and are as such idealized observers. It is important to note, however, that emergence does not lie in the system itself. Nothing would emerge anywhere in the absence of human observers and of their conceptual constructions.

The study of emergence, somewhat unexpectedly, tells us something important about human cognition. The fact that perceptual emergence be possible at all reveals that binary events take place in our mind, and that they occur at different levels. In other words, some aspects of cognition must be structured as detection hierarchies. Furthermore, this is not only true for perception, but also for linguistic processing and conceptual thinking (Bonabeau et al., 1995b). This result should be taken into account in the discussions about the nature of cognitive processing (van Gelder, to appear).

# References

Bennett, C. H. (1990) How to define complexity in physics, and Why, in *Complexity, Entropy and the Physics of Information*, ed. W.H.Zurek, Addison-Wesley.

Bonabeau, E. (1993) Hunting for complexity. Proc. of Second European Conference on Systems Science, pp. 467-475.

Bonabeau, E., Dessalles, J.-L. and Grumbach, A. (1995a) Characterizing emergent phenomena (1): A critical review. *Rev. Int. Syst.* **9**, 327-346.

Bonabeau, E., Dessalles, J.-L. and Grumbach, A. (1995b) Characterizing emergent phenomena (2): a conceptual framework. *Rev. Int. Syst.* **9**, 347-369.

Cariani, P. (1989) *On the design of devices with emergent semantic functions*. Ph.D dissertation, State University of New York at Binghamton.

Cariani, P. (1991) Adaptivity and emergence in organisms and devices, *World Futures* 31, pp. 49-70, Gordon & Breach Science Publishers

Chaitin, G. J. (1979) Toward a mathematical definition of life, in *The maximum entropy formalism*, MIT Press.

Crutchfield, J. and Young, K. (1990) Computation at the onset of chaos, in *Complexity, Entropy and the physics of Information*, ed. W.Zurek, Addison-Wesley.

Dayong, Z. (1990) Detection of spatial pattern in desert shrub populations: a comment. *Ecol. Model.* **51**, 265-271.

Grassberger, P. (1989) Problems in quantifying self-organized complexity. *Helv. Phys. Acta* **62**, 498-511.

Kampis, G. (1991) *Self-Modifying Systems in Biology and Cognitive Science*. Pergamon Press.

Kauffman S. (1990) Requirements for evolvability in complex systems: Orderly dynamics and frozen components, in *Complexity entropy and the physics of information*, Zureck W. H. (ed.), SFI, Addison-Wesley

Lewes, G. H. (1874) in Emergence, *Dictionnaire de la langue philosophique*, Foulquié.

Langton C. (1989) Computations at the edge of chaos, *Physica D, 42*

Nicolis, G. and Prigogine, I. (1989) *Exploring Complexity: An Introduction*. R. Piper, GmbH & Co. KG Verlag.

Palmer R. (1989) Broken ergodicity, in *Complex Systems, SFI Studies in the Sciences of Complexity,* Stein D. (ed.), Addison-Wesley

Pattee H. (1989) in *Artificial Life*, Langton C. (ed.), Addison-Wesley

Rosen, R. (1978) *Fundamentals of measurement and representation of natural systems*, North Holland.

Rosen, R. (1985) *Anticipatory Systems*, Pergamon Press.

Sugihara, G. and May, R. M. (1990) Applications of fractals in ecology. *Trends Eocl. Evol.* **5**, 79-86.

Swindale, N. V. (1980) A model for the formation of ocular dominance stripes. *Proc. Roy. Soc. Lond. B* **208**, 243-264.

Van Gelder T. (to appear) The dynamical hypothesis in cognitive science. *Behavioral and Brain Sciences*

Varela F., Thompson E., Rosch E. (1991) *The embodied mind*, MIT Press.